\documentclass[conference]{IEEEtran}
\IEEEoverridecommandlockouts

\usepackage{array}
\newcommand{\PreserveBackslash}[1]{\let\temp=\\#1\let\\=\temp}
\newcolumntype{C}[1]{>{\PreserveBackslash\centering}p{#1}}
\newcolumntype{R}[1]{>{\PreserveBackslash\raggedleft}p{#1}}
\newcolumntype{L}[1]{>{\PreserveBackslash\raggedright}p{#1}}

\usepackage{float}

\usepackage{caption}
\usepackage{subcaption}

\usepackage{cite}
\usepackage{amsmath,amssymb,amsfonts}
\usepackage{algorithmic}
\usepackage{graphicx}
\usepackage{textcomp}
\usepackage{xcolor}
\def\BibTeX{{\rm B\kern-.05em{\sc i\kern-.025em b}\kern-.08em
    T\kern-.1667em\lower.7ex\hbox{E}\kern-.125emX}}
\begin{document}

\title{Automated Level Crossing System: A Computer Vision Based Approach with Raspberry Pi Microcontroller\\
}
\author{
\IEEEauthorblockN{Rafid Umayer Murshed$^1$, Sandip Kollol Dhruba$^2$, Md. Tawheedul Islam Bhuian$^3$, Mst. Rumi Akter$^4$}
\IEEEauthorblockA{\textit{Department of Electrical and Electronic Engineering} \\
\textit{Bangladesh University of Engineering and Technology, Dhaka-1205, Bangladesh}\\
rafid.buet.eee16@gmail.com$^1$, sandip.kollol@gmail.com$^2$, tawheedrony@gmail.com$^3$, rumimst9701@gmail.com$^4$}
}
\maketitle

\begin{abstract}
In a rapidly flourishing country like Bangladesh, accidents in unmanned level crossings are increasing daily. This study presents a deep learning-based approach for automating level crossing junctions, ensuring maximum safety. Here, we develop a fully automated technique using computer vision on a microcontroller that will reduce and eliminate level-crossing deaths and accidents. A Raspberry Pi microcontroller detects impending trains using computer vision on live video, and the intersection is closed until the incoming train passes unimpeded. Live video activity recognition and object detection algorithms scan the junction 24/7. Self-regulating microcontrollers control the entire process. When persistent unauthorized activity is identified, authorities, such as police and fire brigade, are notified via automated messages and notifications. The microcontroller evaluates live rail-track data, and arrival and departure times to anticipate ETAs, train position, velocity, and track problems to avoid head-on collisions. This proposed scheme reduces level crossing accidents and fatalities at a lower cost than current market solutions.
\end{abstract}


\section{Introduction}
Bangladesh Railway (BR) is a government-owned and operated transportation agency in Bangladesh that covers 2,955.53 route kilometers and employs 25,822 regular employees\cite{Bangladesh_Railway}. According to Bangladesh Railway data, the country has 1,468 authorized and 1,321 unauthorized level crossings. More concerning is that all of the illicit level crossings are unattended, and only 32\% of the authorized level crossings are attended\cite{new_age}. Level crossings which are supposed to ensure the safe passage of trains are currently managed manually. This is understandably unreliable, as many accidents happen due to negligence. Train accidents are nothing new; accidents led to 1,546 recorded casualties over the span of ten years from July 2007 to June 2017, including 365 deaths and 1,181 cases of injury, according to BR\cite{dhaka_tribune_2019}. A solution to end accidents is imperative.

As of now, there have not been many attempts to automate the process. Currently, in all parts of Bangladesh, the system is managed manually by a gate-man who is, after all, a human, and humans make mistakes. This study is conducted to propose a futuristic solution to the ever-increasing problems associated with level crossing junctions, to improve the performance and lessen the casualties across the board. This study is the first of its kind attempting to fully automate the railway level crossing system at minimal realization cost. 

\section{Literature Review}

Many approaches have been taken to solve railway crossing accidents with mixed success. A stereo vision sensor system known as the Ubiquitous Stereo Vision (USV) method, has been tried using stereo cameras that capture 3D images of railway crossings to detect people\cite{hosotani2009development}. Focusing solely on people, it ignores other vehicles that may cross the railway crossing, and since it was tested in Japan, the same rules and conditions don't apply to Bangladesh. Also, the equipment is expensive. Automated video surveillance through object detection and classification using smart background subtraction and Parzen technique\cite{hart2000pattern} considering occlusion of objects, inter-camera travel time, etc. have been tested\cite{shah2007automated}\cite{zhai2006multiple}. In \cite{shah2007automated}, and \cite{zhai2006multiple}, KNIGHT, a Windows-based standalone object detection, tracking, and classification software, has been tested, but it only works during the day, and no testing was done under rain or other adverse weather conditions. Studies on vehicle detection at level crossings by the Histogram of Oriented gradient (HOG) method and the Support Vector Machine (SVM) classifier have been carried out\cite{sugiana2020detection}, but data sets with cars on track, empty track, and trains on tracks were tested separately. Deep learning has been used to first estimate the time of train passing and the number of vehicles under normal traffic conditions in the level crossing; those were used as input to estimate the time of decongestion once the train passes through\cite{jiang2022deep}. This study narrowly focuses on one criterion that concerns the safety of level crossing but does not address the bigger picture. It is no doubt that deep learning algorithms are adept at objection classification tasks, but the challenge is to do so with high accuracy in real time on a mobile device. Exactly this challenge was tackled using deep learning models such as MobileNetSSDLite, Tiny YOLO, etc., to count vehicles and classify them(ATTC) in \cite{sooraj2021real}. As we can see, many techniques have been tried to solve railway crossing problems or other adjacent problems, but to the best of our knowledge, no study adequately tries to detect incoming train and controls the movement of the level crossing barrier accordingly while simultaneously raising the alarm to alert nearby vehicles, humans, etc. This is attempted in our study. 

\section{Methodology}

\begin{figure}[!ht]
    \centering
    \includegraphics[width=\linewidth]{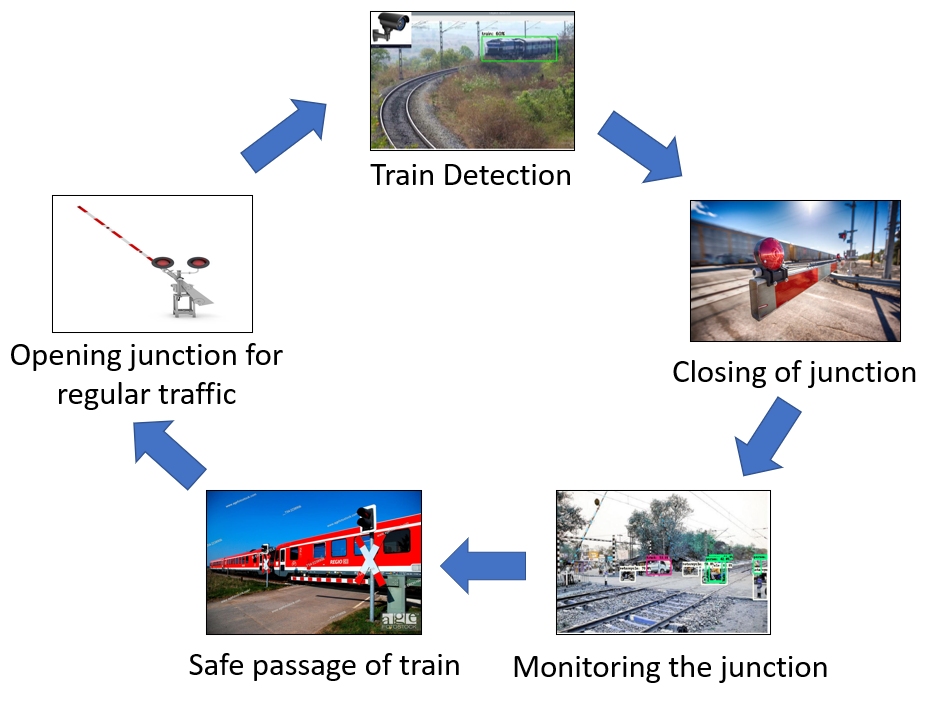}
    \caption{Stages involved in an automated level crossing.}
    \label{Summary of methodology}
\end{figure}

A brief summary of the main tasks completed by the system is shown in Fig. \ref{Summary of methodology}. Briefly, a train is detected, and the junction is closed for safety. Then while the train passes through, the junction is monitored. Finally, the junction is closed after the safe passage of the train. 

\subsection{Train Detection and Junction Closing}
\begin{figure}[!ht]
    \centering
    \includegraphics[width=\linewidth]{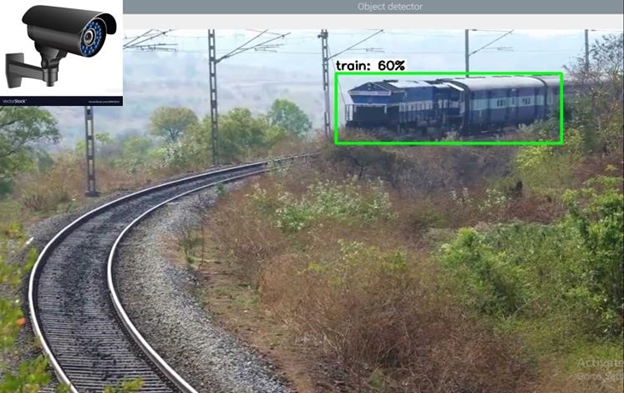}
    \caption{Train detection in normal weather conditions.}
    \label{Train detection in normal conditions}
\end{figure}

\begin{figure}[!ht]
     \centering
     \begin{subfigure}[b]{\linewidth}
         \centering
         \includegraphics[width=\textwidth]{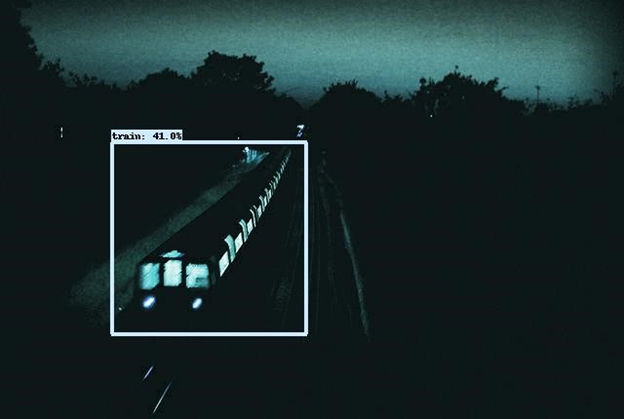}
         \caption{Train detection at night in almost full darkness.}
         \label{Train detection at night}
     \end{subfigure}
     \hfill
     \begin{subfigure}[b]{\linewidth}
         \centering
         \includegraphics[width=\textwidth]{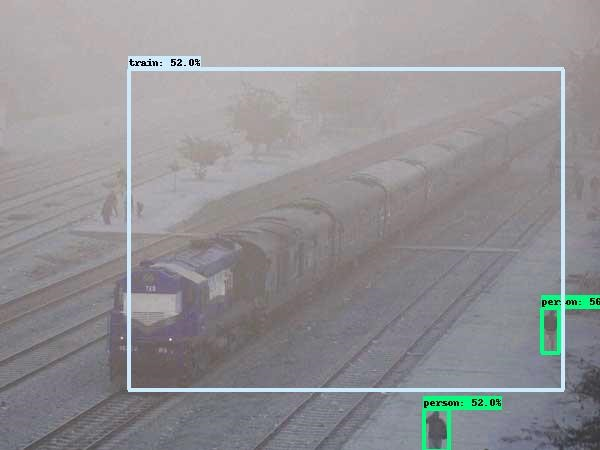}
         \caption{Train detection in foggy condition.}
         \label{Train detection in foggy condition}
     \end{subfigure}
        \caption{Train detection in challenging scenarios due to bad weather conditions.}
        \label{Train detection}
\end{figure}

The main goal of any automated level-crossing system is to detect the presence of an oncoming train. Without that knowledge, no suitable action can be taken. To detect the presence of a train, we rely on several different technologies here. First, we use object detection on a video feed from a camera 2-3 km from the junction as our point of detection. If necessary, this input is prepossessed for object detection. SSD mobilenet v2\cite{chiu2020mobilenet} was trained on the COCO 2017 dataset containing 80 classes and quantized and translated to tflite format for faster inference. This model provides fast inference and low memory needs. Larger and more accurate models can be used based on specific requirements and available computational resources. Upon receiving the camera feed, the Raspberry Pi recognizes the presence of a train and sends a control signal to the peripheral actuators to close the barrier and sound an alarm to inform surrounding pedestrians and traffic. The train's arrival is displayed on a screen or indicator. The junction barrier closes when the Pi sends a control signal, immobilizing nearby vehicles. This is a fully automated process requiring no human input. A CNN-based classification model is also developed to ensure maximum safety and redundancy. Some examples of successful detection are shown in Fig. \ref{Train detection in normal conditions} and \ref{Train detection}.

\begin{figure}[!ht]
    \centering
    \includegraphics[width=\linewidth]{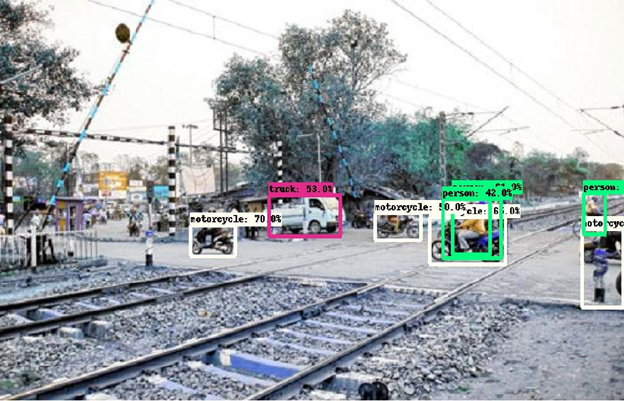}
    \caption{Object detection for activity monitoring at the junction.}
    \label{Object detection for activity monitoring}
\end{figure}

\subsection{Activity Monitoring and Alarms}
No vehicle or pedestrian should cross the intersection once the barrier is closed. A camera at the junction will give a live video feed to the microcontroller, and the object detection model will detect trespassers. If trespassing is detected, the Pi sounds a pre-recorded voice alert and an emergency siren. If the junction isn't cleared after the warnings, the Pi sends a distress signal to neighboring emergency services (police, fire brigade, etc.) for a rescue effort. Also, an emergency signal shall be sent to the incoming train to alert it about possible fatal collisions. Furthermore, an activity identification program may be added to the system to detect abnormal or unusual behavior in the junction zone. Multiple suicides by railway jumping have occurred in Bangladesh in recent decades. To prevent such deplorable acts and mishaps, the junction activity will be constantly monitored by an activity recognition model trained to identify suspicious or unusual behavior. After the train passes, the Pi sends a signal to the actuators and motors to open the junction. Traffic then resumes as usual. An example of successful object detection is shown in Fig. \ref{Object detection for activity monitoring}.

\begin{figure}[!ht]
    \centering
    \includegraphics[width=\linewidth]{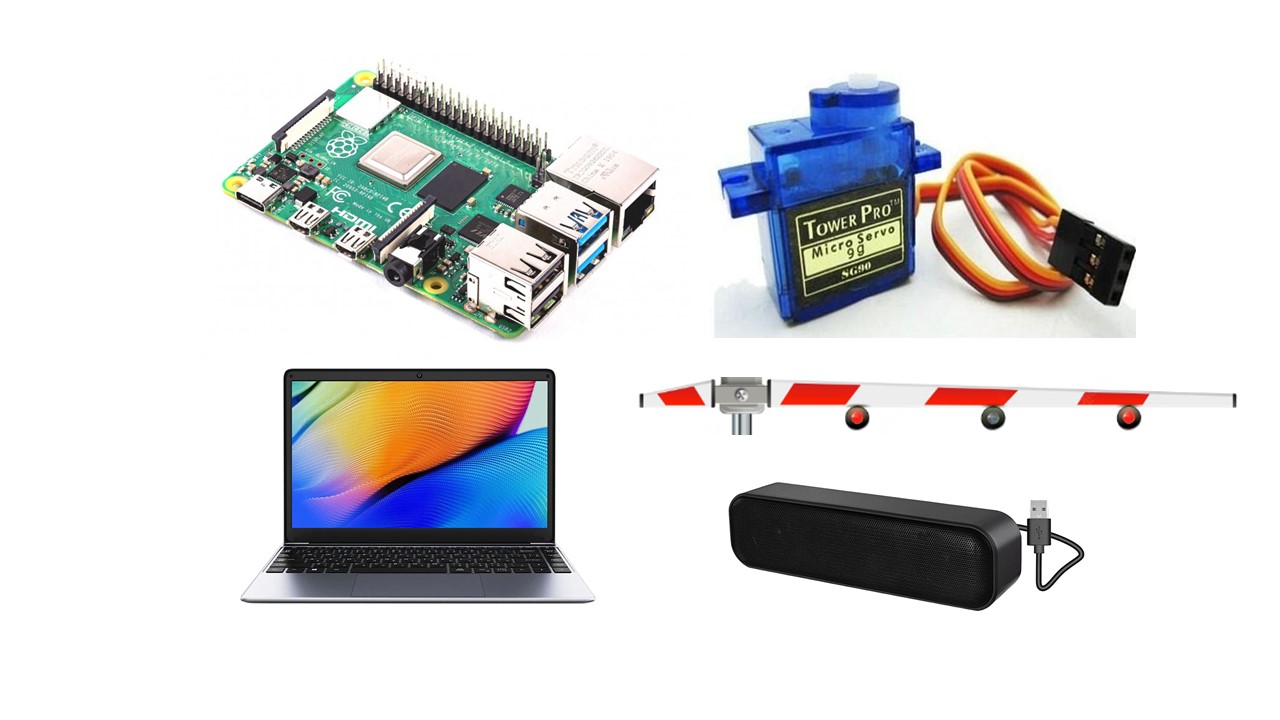}
    \caption{Components used for miniaturized prototype demonstration.}
    \label{Components used for demonstration}
\end{figure}

\section{Experimental Setup and Results}

The goal was to simulate a real-life level-crossing scenario as accurately as possible. The first objective was to detect an incoming train (using the first camera) and bring down a level crossing barrier (blocking traffic flow) while raising the alarm. Once the train leaves, another camera detects that and raises the level crossing barrier again to allow traffic to pass through. While traffic is blocked, the train track is checked for human/vehicle activity (object detection) using a third camera. 

\begin{figure}[!ht]
    \centering
    \includegraphics[width=\linewidth]{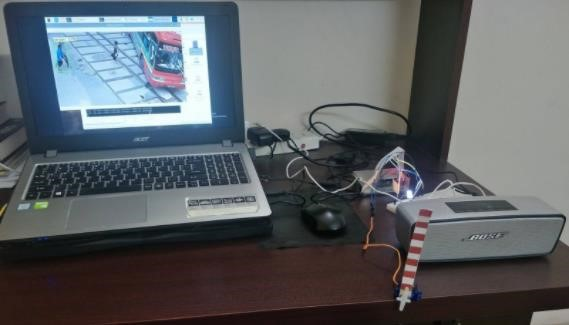}
    \caption{System setup for hardware prototype demonstration.}
    \label{Setup for demonstration}
\end{figure}

\subsection{Components}

The main components used for the demonstration, as shown in Fig. \ref{Components used for demonstration} are Raspberry Pi 4 Model B (4GB RAM), a servo motor (to simulate a larger motor that will lift the level crossing barrier), a speaker (to simulate a buzzer system), a monitor/laptop to view the microcontroller's desktop and a mini level crossing barrier (to simulate a real barrier). The components that would be required for practical implementation include micro-controllers (1 module), camera modules (3 units), copper wires (6km-8km), motor (1 unit), actuator (1 unit), barrier (1 unit), speakers (2 units), sensors (3-8 units) and monitor (1 unit). The estimated cost is around 150k BDT (1400 USD). According to Bangladesh Railway data, the government has spent 1960 million BDT (18.9 million USD) on railway crossings since 2015. From our estimation, the one-time cost of automating the whole process with Raspberry Pi will cost around 420 million BDT (4.05 million USD).

\subsection{Setup \& Data Collection}

\begin{figure}[!ht]
    \centering
    \includegraphics[width=1.05\linewidth]{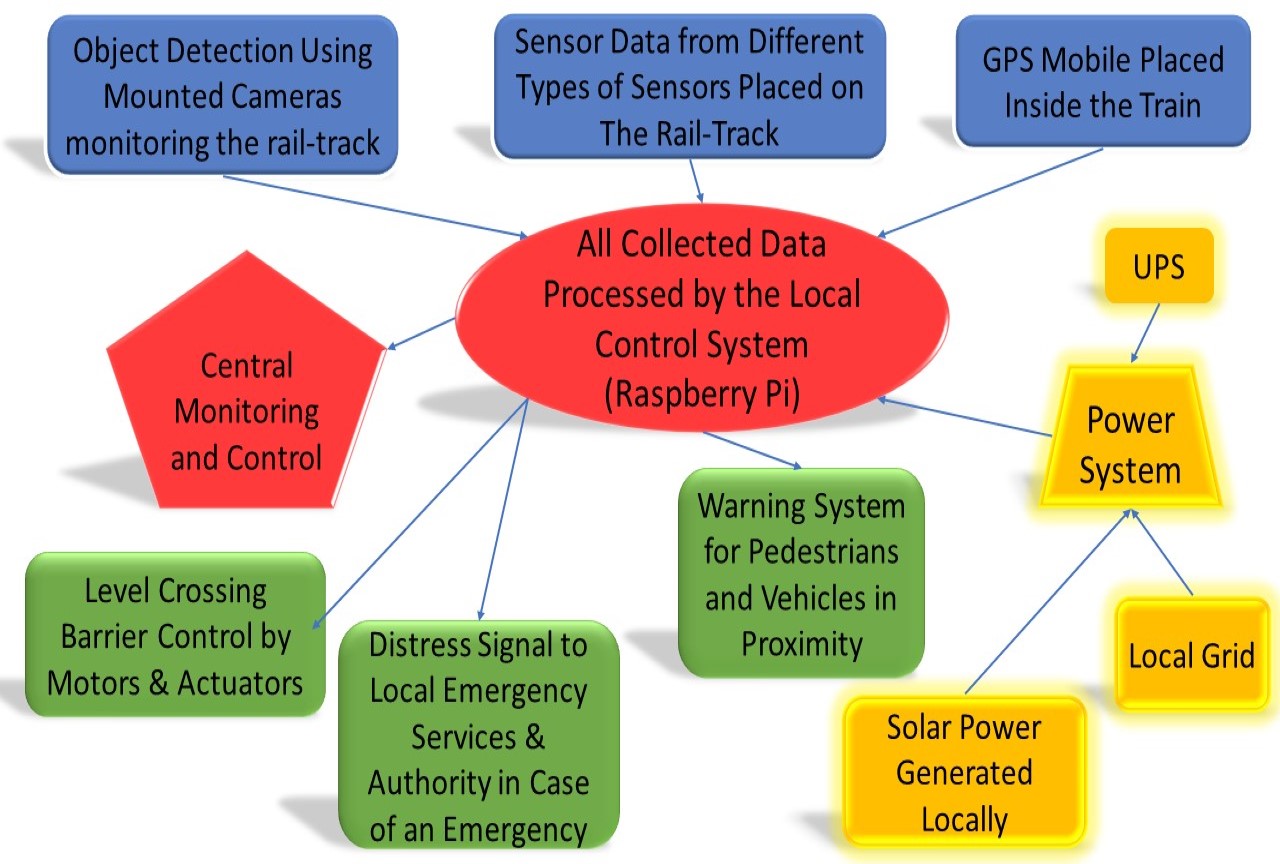}
    \caption{Automated level crossing deployment structure.}
    \label{Automated Level Crossing Flowchart}
\end{figure}

For training and validating our proposed models, data is collected from various well-known computer vision datasets, such as the Image-net dataset, YouTube, and other online platforms. A prototype hardware implementation using a servo motor connected to a Raspberry Pi for demonstration purposes is developed as shown in Fig. \ref{Setup for demonstration}. When a video file containing an event of a train arriving at a level crossing junction is run on the Raspberry Pi for train detection, the train is first detected by the object detection model. Then the Raspberry Pi signals the servo motor to lower the miniature junction barrier, and an alarm is sounded through the connected speaker. Once the train passes by, the junction barrier is again lifted. Although only a few components are used here for demonstration purposes, the proposed real-life system will have a structure similar to the one shown in Fig. \ref{Automated Level Crossing Flowchart}. To increase the robustness and minimize the chance of system failure, the system must be guarded against any power failure. Since load-shedding and complete power outages in Bangladesh are not uncommon, additional power sources must be considered. If power is unavailable at a rail crossing, a generator or solar power arrangement may be necessary, as depicted in Fig. \ref{Automated Level Crossing Flowchart}. 

\subsection{Detection Result}

\begin{table}
\setlength\extrarowheight{10pt}
\caption{Classification model performance analysis.}
\centering
\resizebox{1\linewidth}{!}{%
\begin{tabular}{|C{1.1cm}|C{1cm}|C{0.9cm}|C{0.85cm}|C{0.85cm}|C{0.9cm}|C{0.95cm}|}
\hline
Categories & No. of Images\newline Analyzed & True Positive & True\newline Negative & False\newline Negative & False Positive & Detection\newline Accuracy \\ \hline
Daytime & 8540 & 4582 & 3826 & 93 & 39 & 98.45\%  \\ \hline
Night Time  & 3750  & 2124  & 1247  & 319   & 60  & 89.89\% \\   \hline
Bad Weather & 770  & 329  & 294  & 96  & 51 & 80.91\%  \\ \hline
\end{tabular}
}
\label{Classification Model Output}
\end{table}

\begin{table}
\setlength\extrarowheight{8pt}
\centering
\caption{Object detection model performance analysis.}
\resizebox{1\linewidth}{!}{%
\begin{tabular}{|C{1.05cm}|C{1.1cm}|C{1.1cm}|C{1cm}|C{1.2cm}|C{1.2cm}|}
\hline
Detection\newline Categories &	Number of Frames Analyzed	& Detection Accuracy\newline (Per frame) & False\newline Negatives & Detection Accuracy ( Ten\newline Consecutive Frames ) & False\newline Negatives ( Ten\newline Consecutive Frames ) \\ \hline
Train (Day) & 4533 & 95.6\% & 195 & 100\%  & 0 \\  \hline
Train (Night) & 2158 & 84.3\% & 337 & 99.9\% & 1 \\  \hline
Train (Bad-Weather) & 1326 & 82.7\% & 222 & 99.9\% & 1 \\ \hline
Trespasser (Day) & 3783 & 91.7\% & 306 & 100\%  & 0 \\ \hline
Trespasser (Night) & 2078 & 87.5\% & 269 & 99.9\% & 2 \\ \hline
Trespasser (Bad-Weather) & 1129 & 77.8\% & 234 & 99.7\% & 3 \\ \hline
\end{tabular}
}
\label{Object Detection Model Output}
\end{table}

To verify the performance and robustness of our model, we have tested it with various images ranging from images of trains, humans, cats, dogs, cars, buses, trucks, etc. Fig. \ref{Train detection} shows some extremely challenging images classified and localized perfectly by our model, which shows almost human-level performance. The summarized results of the classification model are shown in Table I. As we can see, the detection accuracy is excellent, considering the diverse conditions tested. The summarized results of the object detection model are also shown in Table II, and again, the number of false negatives is negligible. It is to be noted that the Raspberry Pi microcontroller we used in our prototyping analyzed 5 frames per second on average, making it applicable to real-life critical scenarios.

\section{Conclusion and Future Work}

Unmanned level crossing accidents are rising in a growing country like ours. Our paper concerns the Automatic Level Crossing System, which replaces gatekeepers with automatic junction barrier control. It deals with two things. Firstly it detects the impending train by a camera module positioned at a certain distance. Second, it provides safety by automatically closing and opening the junction barrier. We worked on a foolproof warning system and plan to add more. Our method proposes a cost-effective way to mitigate a fatal flaw in our railroad transport system and reduce casualties. Very-large-scale implementation is possible within a short time due to the wide availability and low cost of the required materials. A pilot project can be conducted with the help of government agencies with minimal effort to prove the effectiveness of the proposed system. If proper support and finance are made available, our proposed method can pave the way for a revolutionary and essential advancement in transport systems safety in developing countries like ours.

\bibliography{bib.bib}
\bibliographystyle{IEEEtran}

\end{document}